\documentclass[lettersize,journal]{IEEEtran}
\usepackage{amsmath,amsfonts}
\usepackage{algorithmic}
\usepackage{algorithm}
\usepackage{array}
\usepackage[caption=false,font=normalsize,labelfont=sf,textfont=sf]{subfig}
\usepackage{textcomp}
\usepackage{stfloats}
\usepackage{url}
\usepackage{float}
\usepackage{verbatim}
\usepackage{graphicx}
\usepackage{cite}
\usepackage{booktabs}
\usepackage{graphicx}
\usepackage{multirow}
\usepackage{lipsum}
\usepackage{amssymb}
\usepackage[utf8]{inputenc}
\usepackage{CJKutf8}
\usepackage{tabularx}
\usepackage{orcidlink}
\usepackage{hyperref}
\usepackage[all]{hypcap}
\hypersetup{
    hypertex=true, 
    colorlinks=true, 
    linkcolor=blue, 
    anchorcolor=blue, 
    citecolor=blue 
}
\begin{document}
\begin{CJK}{UTF8}{gbsn} 
\title{VSD-MOT: End-to-End Multi-Object Tracking in Low-Quality Video Scenes Guided by Visual Semantic
Distillation}


\author{{Jun Du$^{\orcidlink{0009-0007-0383-0631}}$}}



\markboth{}%
{Shell \MakeLowercase{\textit{et al.}}: SE-MOT:End-to-End Multi-Object Tracking in Low-Quality Video
 Scenes Guided by Visual Semantic Enhancement}

\IEEEpubid{}

\maketitle

\begin{abstract}
Existing multi-object tracking algorithms typically fail to adequately address the issues in low-quality videos, resulting in a significant decline in tracking performance when image quality deteriorates in real-world scenarios. This performance degradation is primarily due to the algorithms' inability to effectively tackle the problems caused by information loss in low-quality images. To address the challenges of low-quality video scenarios, inspired by vision-language models, we propose a multi-object tracking framework guided by visual semantic distillation (VSD-MOT). Specifically, we introduce the CLIP Image Encoder to extract global visual semantic information from images to compensate for the loss of information in low-quality images. However, direct integration can substantially impact the efficiency of the multi-object tracking algorithm. Therefore, this paper proposes to extract visual semantic information from images through knowledge distillation. This method adopts a teacher-student learning framework, with the CLIP Image Encoder serving as the teacher model. To enable the student model to acquire the capability of extracting visual semantic information suitable for multi-object tracking tasks from the teacher model, we have designed the Dual-Constraint Semantic Distillation method (DCSD). Furthermore, to address the dynamic variation of frame quality in low-quality videos, we propose the Dynamic Semantic Weight Regulation (DSWR) module, which adaptively allocates fusion weights based on real-time frame quality assessment. Extensive experiments demonstrate the effectiveness and superiority of the proposed method in low-quality video scenarios in the real world. Meanwhile, our method can maintain good performance in conventional scenarios.
\end{abstract}


\section{Introduction}
\IEEEPARstart{M}{ulti-Object} Tracking (MOT) aims to locate and associate multiple objects within video sequences. It has numerous real-world applications, such as autonomous driving\cite{1}, \cite{2}, visual surveillance\cite{3}, \cite{4}, and behavior analysis\cite{5}, \cite{6}. In recent years, although many attempts have been made to improve state-of-the-art MOT techniques\cite{7}, \cite{8}, \cite{9}, \cite{10}, \cite{11},\cite{60}, most of these studies have focused on high-quality video inputs, neglecting the low-quality video scenarios\cite{12} commonly encountered in real-world applications. Based on this, we have investigated multi-object tracking in low-quality video scenarios in the real world.

Existing MOT algorithms for low-quality videos are relatively scarce, and most of them rely on simplified degradation models or assumptions specific to certain scenarios. However, in the real-world scenarios, the task of multi-object tracking is confronted with a multitude of complexities. These include diverse types of noise, non-uniform illumination conditions, intricate backgrounds and occlusions, the diversity and dynamism of targets, varying sensor quality and resolution, changeable environmental conditions, as well as the interactions and occlusions among targets. These complexities pose significant challenges to simplified degradation models and assumptions tailored for specific scenarios, often resulting in suboptimal performance of algorithms based on such assumptions when applied in practical situations. Therefore, developing an MOT algorithm with strong robustness and broad adaptability to handle various low-quality video scenarios is of great significance in both theoretical research and practical applications.
\begin{figure}
    \centering
    \includegraphics[width=1\linewidth]{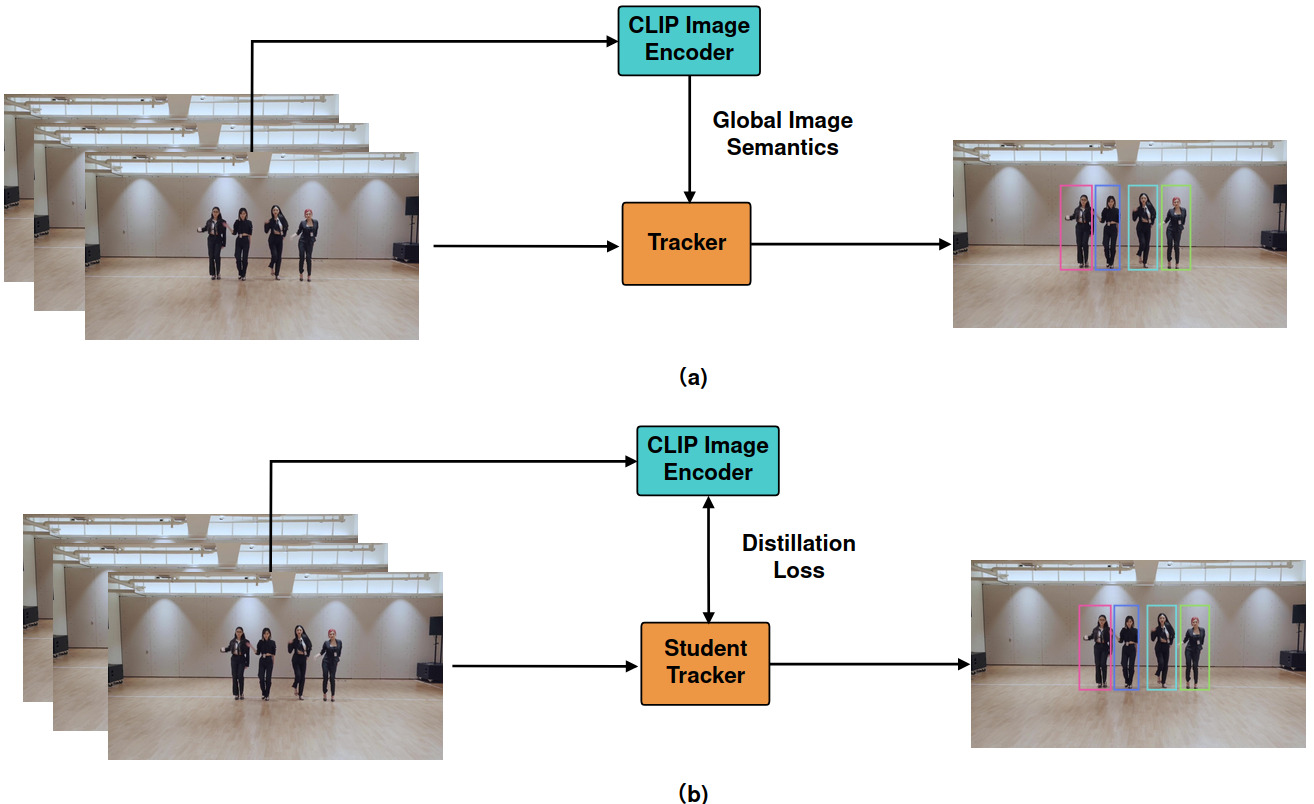}
    \caption{Comparison between (a) the direct integration of the CLIP Image Encoder process and (b) our proposed visual semantic distillation tracking process. The direct integration approach significantly impacts the efficiency of multi-object tracking algorithms. The visual semantic distillation tracking employs a teacher-student learning framework, adding minimal parameters to learn invariant visual semantic information, thereby enhancing tracking performance while ensuring algorithmic efficiency.}
    \label{fig:6}
\end{figure}

Low-quality images typically exhibit significant sensor noise and poor image quality, which severely affect the effectiveness of both shallow and deep feature representations, thereby causing a substantial decline in the performance of multi-object tracking. To address this challenge, our core idea is to learn invariant visual semantic information under the conditions of noise and quality degradation. Inspired by vision-language models, we propose a novel multi-object tracking framework named VSD-MOT. Specifically, we attempt to introduce the CLIP Image Encoder\cite{13} to extract global visual semantic information from images, thereby compensating for the loss of information in low-quality images. However, this direct introduction may have a considerable negative impact on the efficiency of the multi-object tracking algorithm, as depicted in Fig. \ref{fig:6}(a). Therefore, this paper proposes to extract global visual semantic information from images through knowledge distillation\cite{14}, which employs a teacher-student learning framework, as depicted in Fig. \ref{fig:6}(b). Here, the CLIP Image Encoder serves as the teacher model. To adapt the global visual semantic information learned from the teacher model to the multi-object tracking task, this paper proposes the Dual-Constraint Semantic Distillation method(DCSD). Furthermore, considering the dynamic variation of frame quality in low-quality videos—where extremely blurred or high-noise frames have invalid original features while mildly degraded frames retain valuable information—we propose the Dynamic Semantic Weight Regulation (DSWR) module. This module adaptively allocates fusion weights based on real-time frame quality assessment, following the principle of "lower quality, higher semantic weight" to dynamically balance the fusion ratio between visual semantic features and query vector features. The DSWR module operates through three main stages: (1) Frame Quality Assessment, which computes clarity, noise level, and contrast metrics to generate a comprehensive quality score; (2) Weight Generation, which produces adaptive weights through a learnable mapping function based on the quality score; and (3) Adaptive Feature Fusion, which combines visual semantic features and query vector features using the generated weights to ensure optimal tracking performance across varying frame quality conditions. Extensive experiments have demonstrated the effectiveness and superiority of the proposed method in low-quality video scenarios in the real world, while our method maintains good performance in conventional scenarios.

In summary, our main contributions are as follows:
\begin{enumerate}
     \item We propose a novel multi-object tracking method (VSD-MOT) that employs a teacher-student learning framework to learn the ability to extract global visual semantic information from images from the teacher model, the CLIP Image Encoder.
     \item To further improve the utilization of visual semantic information, we propose the DCSD method for efficient knowledge transfer from teacher to student model.
     \item To address the dynamic variation of frame quality in low-quality videos, we propose the Dynamic Semantic Weight Regulation (DSWR) module, which adaptively allocates fusion weights based on real-time frame quality assessment.
     \item We conduct a comprehensive analysis of the proposed method. Experimental results demonstrate the effectiveness and superiority of the proposed method in low-quality video scenarios in the real world, while maintaining good performance in conventional scenarios.
\end{enumerate}

\section{Related Works}
\subsection{Detection-based Tracking Methods}
Detection-based methods dominate the field of multi-object tracking (MOT) and typically follow two main steps: first, object detection, and then, object association between frames. The effectiveness of these methods largely depends on the accuracy of the initial detection stage. Various techniques attempt to utilize the Hungarian algorithm for object association. For example, the SORT\cite{15} algorithm combines the Kalman filter for object state prediction and matches objects by calculating the intersection over union (IoU) between predicted and detected bounding boxes. DeepSORT\cite{16} builds on this by introducing an appearance feature extraction network and optimizes the matching process by calculating the cosine distance. Algorithms such as JDE\cite{17}, TrackRCNN\cite{18}, FairMOT\cite{19}, and Unicorn\cite{20} further explore the joint optimization of detection and appearance features. ByteTrack\cite{21} employs an advanced detector based on YOLOX\cite{22} and incorporates low-confidence detection results into the association process by improving the SORT algorithm. BoT-SORT\cite{23} enhances performance by optimizing Kalman filter parameters, compensating for camera motion, and integrating ReID features. TransMOT\cite{24} and GTR\cite{25} utilize spatio-temporal transformers to process instance features and aggregate historical information to form an assignment matrix. OC-SORT\cite{26} abandons the strict linear motion assumption and instead adopts a learnable dynamic model.

\subsection{Tracking by Query Propagation}
Another paradigm in multi-object tracking (MOT) extends query-based object detectors\cite{27}, \cite{28}, \cite{29} to tracking. These methods track the same instances by maintaining the consistency of queries across different frames. The interaction between queries and image features can be carried out either in parallel or serially.

Parallel methods process short video clips by allowing a set of queries to interact with the features of all frames to predict trajectories. VisTR\cite{30} and its derivative methods\cite{31}, \cite{32} apply DETR\cite{27} for trajectory detection in short video clips. These methods consume a large amount of memory as they need to process the entire video and can only handle short video clips.

Sequential methods update queries frame by frame, iteratively optimizing the tracking results through interactions with image features. Tracktor++\cite{33} iteratively updates the instance positions using the regression head of R-CNN\cite{34}. TrackFormer\cite{35} and MOTR\cite{36} extend Deformable DETR\cite{37} to predict bounding boxes and update queries for continuous tracking. MeMOT\cite{38} constructs short-term and long-term feature memory banks to generate tracking queries. TransTrack\cite{39} predicts the positions of objects in the next frame through single-query propagation. P3AFormer\cite{40} utilizes optical flow-guided feature propagation. Different from MOTR, TransTrack and P3AFormer still use position-based Hungarian matching when dealing with historical trajectories and current detections instead of propagating queries across the entire video sequence. MOTRv2\cite{41} uses YOLOX as an object detector to generate high-quality object proposals, which are input as anchors into an improved MOTR tracker to learn tracking associations. MOTIP\cite{42} treats the object association task as a context ID prediction problem.

\begin{figure*}
    \centering
    \includegraphics[width=1\linewidth]{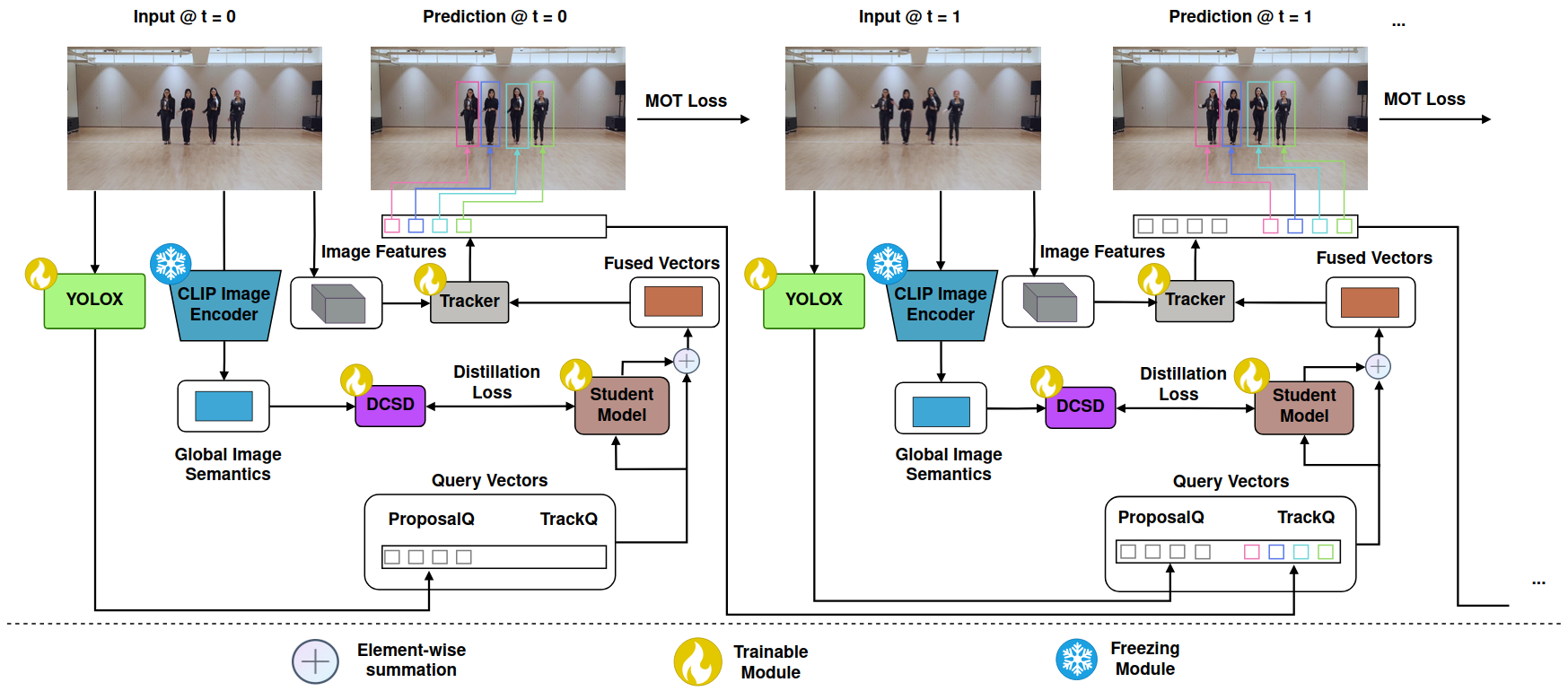}
    \caption{The overall architecture of VSD-MOT. The frozen CLIP Image Encoder extracts global visual semantic information from images, and proposals generated by the  detector YOLOX are used to produce proposal queries. Tracking inquiries are transferred from the preceding frame for the purpose of forecasting the bounding boxes of tracked objects. The combination of proposal queries and track queries generates query vectors. Through the Dual-Constraint Semantic Distillation (DCSD) method, the student model can learn from the CLIP Image Encoder the capability to extract semantic information that is adaptive to multi-object tracking tasks. The semantic information extracted by the student model is integrated with query vectors through the Dynamic Semantic Weight Regulation (DSWR) module to generate adaptively fused vectors. Both the fused vectors and the image features are fed into the tracker to produce predictions frame by frame.}
    \label{fig:1}
\end{figure*}

\subsection{Vision-language Models}
Vision-language models have been extensively studied in fields such as text-to-image retrieval, image captioning, visual question answering, and referential segmentation. Recently, vision-language pre-training has gained increasing attention, with the landmark work being Contrastive Language-Image Pretraining (CLIP). CLIP pre-trains the model through contrastive learning on 4 million image-text pairs crawled from the Internet. It demonstrates impressive generalization ability in evaluations on 30 classification datasets. The pre-trained CLIP encoder has also been applied to many other downstream tasks, such as open-vocabulary detection and zero-shot visual semantic segmentation. Recently, some follow-up works have attempted to leverage pre-trained models to deal with the video domain. For example, CLIP4Clip\cite{43} transfers the knowledge of the CLIP model to video-text retrieval, and Actionclip\cite{44} attempts to utilize the capabilities of CLIP for video recognition. In addition, CLIP has been used to address the challenges of complex video action localization (Nag et al.)\cite{45}.

·\subsection{Knowledge Distillation}
Knowledge distillation is a model compression technique that enhances the performance of smaller models (student models) by transferring knowledge from larger models (teacher models) via distillation loss\cite{46}, \cite{47}, \cite{48}. In recent years, as the scale and number of parameters of deep learning models have continuously increased, the application of knowledge distillation in the field of multi-object tracking (MOT) has become increasingly significant.

Summarizing the above tracking methods, they lack consideration for low-quality video scenarios. Features such as blurring, noise, and uneven illumination in low-quality videos make it difficult for traditional tracking algorithms to work effectively. Our method leverages the rich visual semantic information from the CLIP model to further classify and associate objects in low-quality videos. However, a direct introduction would have a relatively large impact on the efficiency of multi-object tracking algorithms. Therefore, this paper proposes to extract the global visual semantic information of images from the CLIP Image Encoder through knowledge distillation.

\section{Method}
In this work, we propose a multi-object tracking framework named VSD-MOT, the overall architecture of which is illustrated in Fig. \ref{fig:1}. Within the proposed framework, first, we adopt a teacher-student learning framework to learn the ability to extract visual semantic information from the teacher model, the CLIP Image Encoder. Second, to enable more efficient transfer of the semantic extraction capability from the CLIP Image Encoder to the student model, we propose the Dual-Constrained Semantic Distillation method (DCSD), which leverages two complementary loss terms: a local feature matching loss and a global feature alignment loss.

\begin{figure*}
    \centering
    \includegraphics[width=1\linewidth]{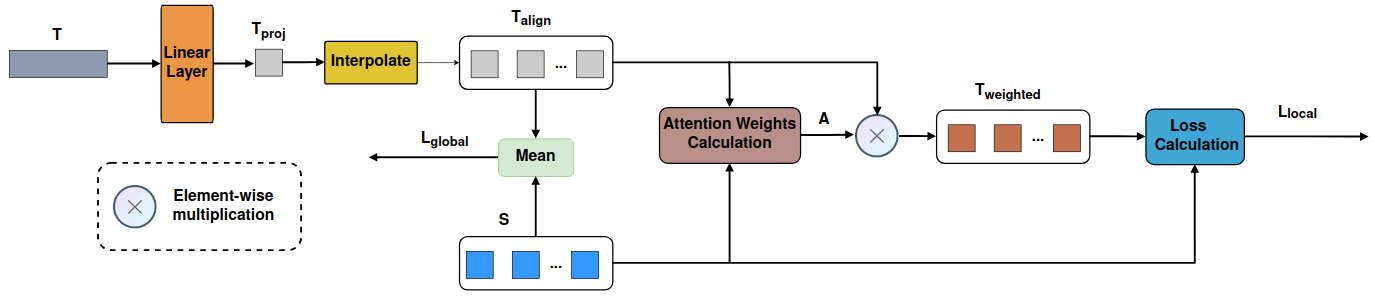}
    \caption{Structure of Dual-Constrained Semantic Distillation. To enable the student model to better extract semantic information adapted to the multi-object tracking task, the core of this module implements a dual-constraint mechanism via two complementary losses, which balances local feature matching and global semantic consistency. }
    \label{fig:3}
\end{figure*}

\subsection{Student Model}
\begin{figure}
    \centering
    \includegraphics[width=1\linewidth]{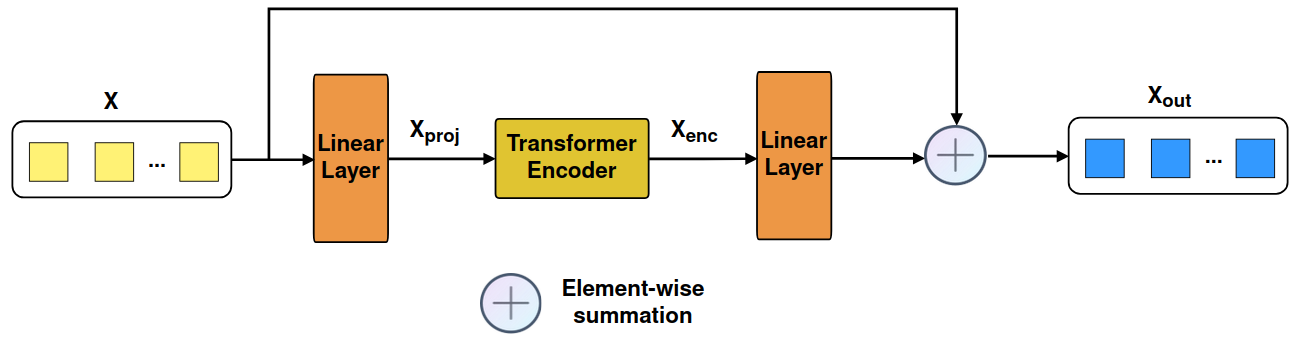}
    \caption{Structure of student model. It employs a Transformer Encoder architecture to process the query vectors of the multi-object tracking algorithm.}
    \label{fig:2}
\end{figure}

The structure of the student model in this paper is shown in Fig. \ref{fig:2}.  \(X \in \mathbb{R}^{n \times 256}\) represents the input sequence features, i.e., the query vectors of the multi-object tracking algorithm, where n is the length of the input sequence and 256 is the dimension of the input features. First, it is necessary to project the input features to a suitable hidden dimension. The input feature map X is processed through a linear projection layer to adjust the feature dimension, resulting in the feature map \(X_{proj} \in \mathbb{R}^{n \times hidden\_dim}\), which is expressed as:
\begin{equation}
\label{deqn_ex1a}
X_{proj} = \text{Linear}(X)
\end{equation}
where Linear(·) represents the linear projection layer.
Subsequently, the projected features are fed into the Transformer Encoder to capture sequence relationships while keeping the sequence length unchanged. This process generates the encoded feature map \(X_{enc} \in \mathbb{R}^{n \times hidden\_dim}\), which can be expressed as:
\begin{equation}
\label{deqn_ex1a}
X_{enc} = \text{TransformerEncoder}(X_{proj})
\end{equation}
where TransformerEncoder(·) denotes the Transformer Encoder module composed of 3 Transformer Encoder layers.
To obtain the final output features with the target dimension, the encoded features undergo another linear projection. Meanwhile, a residual connection is introduced to enhance feature propagation and training stability. The output feature map \(X_{\text{out}} \in \mathbb{R}^{n \times 256}\) is expressed as:
\begin{equation}
\label{deqn_ex1a}
X_{out} = \text{Linear}(X_{enc}) + \text{Residual}(X)
\end{equation}
where Linear(·) is the output projection layer, and Residual(·) represents the residual connection layer.

\subsection{Dual-Constraint Semantic Distillation}
To better adapt the global visual semantic information of images to multi-object tracking tasks, we propose the Dual-Constrained Semantic Distillation method (DCSD). Fig. \ref{fig:3} illustrates the structure of the proposed DCSD method. \(S \in \mathbb{R}^{n \times 256}\) represents the student output features, and \(T \in \mathbb{R}^{1 \times 1024}\) denotes the teacher output features. To align the teacher features with the output dimension of the student model, the teacher features undergo linear projection, resulting in the feature map \(T_{proj}\in \mathbb{R}^{1 \times 256}\), represented as:
\begin{equation}
\label{deqn_ex1a}
T_{proj} = \text{Linear}_{teacher}(T) 
\end{equation}
where \(\text{Linear}_{teacher}(\cdot)\) is a linear layer that maps the teacher features from 1024 dimensions to 256 dimensions. The projected teacher features are aligned to a length of n via linear interpolation, resulting in the feature map \(T_{align}\in \mathbb{R}^{n \times 256}\), represented as:
\begin{equation}
\label{deqn_ex1a}
T_{align} = \text{Interpolate}(T_{proj})
\end{equation}
where \(\text{Interpolate}(\cdot)\) adjusts the sequence length while preserving the feature dimension.

To establish fine-grained associations between the student and teacher features, normalized attention weights are computed, resulting in the feature map \(A\in \mathbb{R}^{n \times n}\), represented as:
\begin{equation}
\label{deqn_ex1a}
S_{norm} = \text{Normalize}(S) 
\end{equation}
\begin{equation}
\label{deqn_ex1a}
T_{norm} = \text{Normalize}(T_{align}) 
\end{equation}
\begin{equation}
\label{deqn_ex1a}
A = \text{Softmax}\left( \frac{S_{norm} \cdot T_{norm}^T}{\tau} \right) 
\end{equation}
where \(\text{Normalize}(\cdot)\) refers to L2 normalization, \(\tau\) is a temperature parameter (set to 2.0), and A represents the attention scores between positions in the student and teacher sequences. The teacher features are then aggregated using these attention weights, resulting in the feature map \(T_{weighted}\in \mathbb{R}^{n \times 256}\), represented as:
\begin{equation}
\label{deqn_ex1a}
T_{weighted} = A \cdot T_{align} 
\end{equation}

The distillation loss calculation employs two complementary loss terms to align student and teacher features. The local feature matching loss measures the feature similarity at each position, represented as:
\begin{equation}
\label{deqn_ex1a}
\mathcal{L}_{\text{local}} = \text{MSE}(S, T_{\text{weighted}})
\end{equation}
where \(\text{MSE}(\cdot)\) calculates the mean squared error between two feature tensors. The global feature alignment loss ensures the consistency of sequence-level statistics, represented as:
\begin{equation}
\label{deqn_ex1a}
\mathcal{L}_{\text{global}} = \text{L1}\left( \text{Mean}(S), \text{Mean}(T_{\text{align}}) \right)
\end{equation}
where \(\text{L1}(\cdot)\) computes the L1 loss (mean absolute error) between two feature vectors, and \(\text{Mean}(\cdot)\) denotes the calculation of the mean value of a feature sequence. The final distillation loss is weighted and combined using learnable weights \(w_1, w_2\) normalized by softmax, which is expressed as:
\begin{equation}
\label{deqn_ex1a}
\mathcal{L}_{\text{distill}} = w_1 \cdot \mathcal{L}_{\text{local}} + w_2 \cdot \mathcal{L}_{\text{global}}
\end{equation}

\begin{figure}[h]
\centering
\includegraphics[width=0.95\linewidth]{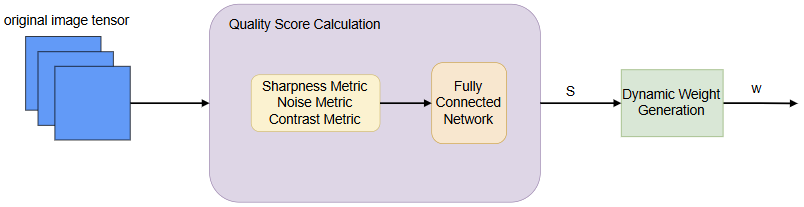}
\caption{Network architecture of Dynamic Semantic Weight Regulation Module. The module evaluates frame quality through clarity, noise level, and contrast metrics, then generates adaptive fusion weights for combining visual semantic features with query vector features.}
\label{fig:dswr}
\end{figure}

\subsection{Dynamic Semantic Weight Regulation Module}
Low-quality videos commonly exhibit the phenomenon of "frame quality fluctuation"—extremely blurred or high-noise frames tend to have invalid original visual features, while mildly degraded or normal frames still retain valuable original information. Fixed-ratio fusion of visual semantic features and query vector features fails to adapt to this dynamic variation. For extremely low-quality frames, fixed ratios result in excessive invalid original features, preventing visual semantic features from effectively compensating for information loss. For mildly degraded or normal frames, over-reliance on semantic features wastes effective original information, ultimately reducing tracking accuracy stability.

To address this issue, we propose the Dynamic Semantic Weight Regulation (DSWR) module, which adaptively allocates fusion weights based on real-time frame quality assessment. Following the principle of "lower quality, higher semantic weight," this module dynamically balances the fusion ratio between visual semantic features and query vector features, thereby enhancing cross-scene robustness and stability.

As illustrated in Fig. \ref{fig:dswr}, the DSWR module operates through three main stages:

\textbf{Frame Quality Assessment:} The module computes three key metrics from the grayscale frame: clarity (using Laplacian variance), noise level (through high-frequency component analysis), and contrast (via standard deviation measurement). These metrics are normalized and combined into a comprehensive quality score \(Q \in [0, 1]\), where lower values indicate poorer frame quality.

\textbf{Weight Generation:} Based on the quality score \(Q\), the module generates adaptive weights through a learnable mapping function:
\begin{equation}
\label{eq:weight_gen}
w_{\text{semantic}} = \sigma(W \cdot Q + b)
\end{equation}
where \(W\) and \(b\) are learnable parameters, and \(\sigma(\cdot)\) is the sigmoid function ensuring weights fall within \((0, 1)\). The semantic weight \(w_{\text{semantic}}\) increases as frame quality decreases.

\textbf{Adaptive Feature Fusion:} The final fused feature \(F_{\text{fused}}\) combines visual semantic features \(F_{\text{semantic}}\) and query vector features \(F_{\text{query}}\) using the generated weights:
\begin{equation}
\label{eq:adaptive_fusion}
F_{\text{fused}} = w_{\text{semantic}} \cdot F_{\text{semantic}} + (1 - w_{\text{semantic}}) \cdot F_{\text{query}}
\end{equation}

This adaptive mechanism ensures that for high-quality frames, the model relies more on original query features while maintaining semantic guidance, whereas for low-quality frames, it increasingly leverages semantic information to compensate for degraded visual cues. The DSWR module operates with minimal computational overhead, adding negligible parameters while significantly improving tracking stability across varying frame quality conditions.

\subsection{Loss Functions}
As illustrated in Fig. \ref{fig:1}, the frozen CLIP Image Encoder extracts global visual semantic information from the images. Enable the student model to better extract semantic information suitable for multi-object tracking tasks through the DCSD method. Consequently, the losses for the detection and re-identification heads are calculated in the conventional manner, and the distillation loss should be also incorporated into the total loss. The final loss function is formulated as follows:
\begin{equation}
\label{deqn_ex1a}
loss= \alpha  \mathcal{L}_{\text{distill}} + (1 - \alpha)  \cdot \mathcal{L}_{\text{mot}}
\end{equation}
where \(\mathcal{L}_{\text{distill}}\) is the distillation loss, \(\mathcal{L}_{\text{mot}}\) is the conventional MOTRv2 loss function, and \(\alpha\) is a hyperparameter within the range \((0, 1)\) that determines the proportion of the distillation loss added to the overall loss function.
\begin{equation}
\label{deqn_ex1a}
loss= w  \mathcal{L}_{\text{distill}} + (1 - w)  \cdot \mathcal{L}_{\text{mot}}
\end{equation}
\section{Experiments}
To confirm both the advantages of the proposed VSD-MOT in low-quality video scenarios and its ability to sustain stable performance under conventional conditions, two dedicated low-quality multi-object tracking datasets were developed. Specifically, these datasets include: (1) LQDanceTrack, which is derived from the original DanceTrack dataset; and (2) LQMOT, built based on the MOT17 and MOT20 datasets.​

In this section, the experimental outcomes of the proposed VSD-MOT are presented, and a comparative analysis is conducted between this method and other cutting-edge approaches. The comparison is carried out on the validation subsets of four datasets: the standard DanceTrack dataset, the low-quality LQDanceTrack dataset, the conventional MOT dataset, and the low-quality LQMOT dataset.​

Furthermore, to conduct a thorough evaluation and verification of the performance of each module included in the proposed method, a series of comprehensive ablation experiments were designed. The results of these experiments indicate that the proposed VSD-MOT not only achieves effective and superior performance in real-world low-quality video environments but also maintains strong performance when applied to traditional video scenarios.

\subsection{Dataset}
{\bf{DanceTrack:}} DanceTrack\cite{51} is a large-scale multi-person tracking dataset designed for dance scenarios. It exhibits consistent visual traits alongside varied movement styles, a combination that brings about difficulties in associating instance objects across different video frames. The DanceTrack dataset is made up of 100 video clips, divided into three subsets: 40 clips are designated for training purposes, 25 for validation, and 35 for testing. On average, each video in this dataset lasts 52.9 seconds.

{\bf{MOT17:}} MOT17 \cite{61} is a commonly adopted dataset that includes 7 sequences for training and another 7 for testing. Its main characteristic lies in relatively congested street settings, where people on foot display movement that is straightforward and follows a linear pattern.

{\bf{MOT20:}} The MOT20 \cite{62} dataset comprises 8 video sequences, with 4 allocated to the training set and 4 to the test set, focusing on high-density pedestrian tracking in complex environments.

{\bf{LQDanceTrack:}} To advance research in the field of multi-object tracking—especially when dealing with low-quality video inputs—we took cues from recent research efforts that replicate degraded scenarios encountered in real-world environments. In line with this inspiration, we constructed a low-quality dataset titled LQDanceTrack. Originating from the existing DanceTrack dataset, this new dataset integrates the characteristic traits of real-world low-quality videos by leveraging the degradation function of the Real-ESRGAN model\cite{52}. In detail, we randomly chose two-thirds of the video sequences from the training subset of DanceTrack, and then applied the degradation function of Real-ESRGAN to these selected sequences, thereby simulating low-quality video conditions. The degradation function is formulated as follows:
\begin{equation}
\label{deqn_ex1a}
x=D_{n}(y)=(D_{n}\circ{\bf\cdot}\cdot{\bf\cdot}\circ D_{2}\circ D_{1})(y)
\end{equation}
where \textit{D$_1$},\textit{D$_2$},…,\textit{D$_n$ }represent a sequence of degradation operators applied successively to the original image \textit{y}, where each \textit{D$_i$ }corresponds to a specific type of degradation, such as blurring, downsampling, or noise addition.  $\circ $ denotes the functional composition, meaning the sequential execution of functions where the output of one function serves as the input to the next, \textit{D$_n$}(\textit{y}) indicates the final output after all degradation operations have been applied. Furthermore, we extended the same degradation approach to the validation subset of DanceTrack, leveraging it to construct the corresponding validation set for our low-quality dataset, LQDanceTrack.

{\bf{MOT, LQMOT:}} We curated the MOT training set by choosing 5 video sequences from the MOT17 training set and 2 video sequences from the MOT20 training set; meanwhile, the leftover portions of the MOT17 and MOT20 training sets were used to form the MOT validation set. After that, we adopted the same construction strategy that was employed for building the LQDanceTrack training and validation sets, and applied this approach to process the MOT training and validation sets—through this process, we successfully obtained the LQMOT training set and LQMOT validation set.

{\bf{Mixed Training Sets:}} To allow the multi-object tracking model to adapt to both real-world low-quality video scenarios and traditional video scenarios, we merged two pairs of datasets respectively: on one hand, we combined the LQDanceTrack training set with the undegraded DanceTrack training set; on the other hand, we merged the LQMOT training set with the undegraded MOT training set. Through this process, we constructed two mixed training sets, where the proportion of low-quality videos to high-quality videos is maintained at 2:1.​ We then conducted training on these two mixed training sets separately—specifically, we trained the VSD-MOT model as well as other state-of-the-art methods. After training, we performed testing on four validation sets individually: the LQDanceTrack validation set, the DanceTrack validation set, the LQMOT validation set, and the MOT validation set. This experimental design is intended to verify two key points: first, the effectiveness and superiority of VSD-MOT in real-world low-quality video scenarios; second, whether VSD-MOT can retain good performance in conventional video scenarios. In addition, we also carried out ablation experiments on the LQMOT validation set and the MOT validation set.

\subsection{Evaluation Metrics}
To evaluate our method and dissect its contributions, we utilize Higher Order Tracking Accuracy (HOTA)\cite{53}, which is further broken down into Detection Accuracy (DetA) and Association Accuracy (AssA). Additionally, we include Multi-Object Tracking Accuracy (MOTA)\cite{54} and Identity Filter Precision (IDF1)\cite{55} as additional evaluation metrics.

\subsection{Experimental Details}
All experiments were conducted using the Python programming language and the PyTorch framework. The training phase was performed on a device equipped with 8 NVIDIA GeForce RTX 4090 GPUs. The inference was executed on a single NVIDIA GeForce RTX 4090 GPU. The generation of target candidate regions employed the YOLOX detection algorithm, and to maximize the recall rate of candidate regions, all YOLOX predictions with a confidence score exceeding 0.05 were retained as candidate regions. For the hybrid training set of LQDanceTrack and DanceTrack, the YOLOX detector was trained for 20 epochs on 8 GPUs. For the hybrid training set of LQMOT and MOT, the detector was trained for 50 epochs under the same 8-GPU configuration. 

We employed ResNet50\cite{56} as the backbone network for feature extraction. The model’s entire training process was carried out using 8 GPUs, with each GPU configured to handle a batch size of 1. When training VSD-MOT on the combined dataset of LQMOT and MOT, we incorporated HSV augmentation techniques, aligning with the established methodology employed in the YOLOX framework. To strengthen the model’s ability to address false positives (FPs) and false negatives (FNs) during the inference phase, we transmitted tracking queries that had confidence scores exceeding 0.5. This approach inherently produces two types of problematic tracking queries: FPs (queries with high confidence scores but corresponding to no actual instances, like incomplete trajectories) and FNs (queries that fail to detect existing instances).In both the ablation experiments and the comparative analyses against state-of-the-art models, the training was conducted for 60 epochs, and a fixed segment size of 5 was maintained throughout. Within each segment, the sampling step was randomly chosen from the range of 1 to 10. The initial learning rate was set to \(2 \times 10^{-4}\), and it was scaled down by a factor of 10 starting from the \(40^{th}\) epoch.For the mixed training set consisting of LQDanceTrack and DanceTrack, adjustments were made to the training parameters: the total number of training epochs was decreased to 10, and the learning rate was reduced at the \(5^{th}\) epoch. 

To boost the model’s detection performance, a substantial number of static images from the CrowdHuman dataset were utilized. Specifically, when training on the LQDanceTrack-DanceTrack mixed set, joint training was performed using both the train and validation sets of CrowdHuman; in contrast, for the LQMOT-MOT mixed set, joint training only leveraged the validation set of CrowdHuman.
\begin{table}[htbp]
\centering
\caption{Performance Comparison of Different Trackers on the LQDanceTrack Validation Set. ↑ Indicates Higher Is Better, the Best Results for Each Metric Are Highlighted in Bold. The Proposed
VSD-MOT Outweighs Other Methods on the LQDanceTrack
Validation Set and Achieves the Best in All Metrics, Demonstrating the Effectiveness of the Proposed Method in Real-world Low-quality Video Scenarios}
 \label{tab:1}
\Huge
\resizebox{\columnwidth}{!}{%
\begin{tabular}{c|ccccc}
\toprule
Tracker & HOTA(\%)↑ & DetA(\%)↑ & AssA(\%)↑ & MOTA(\%)↑ & IDF1(\%)↑ \\
\midrule
CO-MOT \cite{58} & 42.0 & 54.4 & 32.9 & 53.3 & 43.0 \\
MeMOTR \cite{59} & 49.1 & 58.1 & 42.1 & 63.7 & 52.0 \\
MOTIP \cite{39} & 52.2 & 61.2 & 45.3 & 67.8 & 57.1 \\
Hybrid-SORT-ReID \cite{60} & 41.8 & 52.3 & 33.7 & 59.8 & 44.3 \\
Hybrid-SORT \cite{60} & 37.3 & 52.4 & 26.8 & 59.6 & 38.9 \\
MOTR \cite{34} & 39.9 & 51.8 & 31.3 & 46.3 & 41.4 \\
MOTRv2(baseline) \cite{11} & 51.4 & 67.3 & 39.6 & 65.2 & 53.2 \\
\hline
VSD-MOT(ours)& \textbf{59.7}& \textbf{73.6}& \textbf{48.7}& \textbf{82.6}& \textbf{61.2}\\
\bottomrule
\end{tabular}%
}
\end{table}

\begin{table}[htbp]
\centering
\caption{Performance Comparison of Different Trackers on the DanceTrack Validation Set. ↑ Indicates Higher Is Better, the Best Results for Each Metric Are Highlighted in Bold. The Proposed VSD-MOT Outperforms Other Methods
on the DanceTrack Validation Set and Achieves the Best in All Metrics. It Shows That after Training on the Mixed Training Set, the Proposed Method Can Maintain Good
Performance in Conventional Scenarios}
\label{tab:2}
\Huge
\resizebox{\columnwidth}{!}{%
\begin{tabular}{c|ccccc}
\toprule
Tracker & HOTA(\%)↑ & DetA(\%)↑ & AssA(\%)↑ & MOTA(\%)↑ & IDF1(\%)↑ \\
\midrule
CO-MOT \cite{58}& 46.7& 54.8& 40.5& 44.9& 47.3\\
MeMOTR \cite{59}&  59.3& 69.9& 50.7& 78.1& 62.7\\
MOTIP \cite{39}& 62.1& 73.5& \textbf{52.8}& 83.8& \textbf{66.9}\\
Hybrid-SORT-ReID \cite{60}& 49.3& 56.5& 43.5& 65.3& 47.4\\
Hybrid-SORT \cite{60}& 44.5& 56.3& 35.6& 64.5& 43.2\\
MOTR \cite{34}& 48.4& 59.2& 39.8& 57.5& 50.2\\
MOTRv2(baseline) \cite{11}& 58.4& 69.9& 49.0& 80.2& 61.4\\
\hline
VSD-MOT(ours)& \textbf{63.4}& \textbf{78.2}& 51.6& \textbf{87.2}& 64.0\\
\bottomrule
\end{tabular}%
}
\end{table}
\subsection{Performance Comparison with the State-of-the-Art Methods on LQDanceTrack and DanceTrack}
To showcase the efficacy of our method in real-world low-quality video scenarios and its capacity to sustain robust performance in conventional scenarios, we conducted training of VSD-MOT and other state-of-the-art methods on a mixed training set comprising low-quality and high-quality videos in a 2:1 ratio. Subsequently, we evaluated these methods on the validation sets of LQDanceTrack and DanceTrack.  The results are detailed in Table \ref{tab:1} and \ref{tab:2}, with the top-performing metrics highlighted in boldface. As illustrated in Table \ref{tab:1}, the proposed VSD-MOT outperforms other methods on the LQDanceTrack validation set, achieving the highest scores across all metrics. Compared to other state-of-the-art methods, VSD-MOT demonstrates a 8\% to 20\% lead in each metric, underscoring its effectiveness in real-world low-quality video scenarios. As shown in Table \ref{tab:2}, the proposed VSD-MOT also achieves superior performance on the DanceTrack validation set, with the best results in all metrics. Compared to other state-of-the-art methods, it shows a 8\% to 20\% lead. These findings indicate that after training on the mixed dataset, our proposed method maintains strong performance in conventional scenarios.

\begin{table}[htbp]
\centering
\caption{Performance Comparison of Different Trackers on the LQMOT Validation Set. ↑ Indicates Higher Is Better, the Best Results for Each Metric Are Highlighted in Bold. The Proposed
VSD-MOT Outweighs Other Methods on the LQMOT
Validation Set, Demonstrating the Effectiveness of the Proposed Method in Real-world Low-quality Video Scenarios.}
 \label{tab:3}
\Huge
\resizebox{\columnwidth}{!}{%
\begin{tabular}{c|ccccc}
\toprule
Tracker & HOTA(\%)↑ & DetA(\%)↑ & AssA(\%)↑ & MOTA(\%)↑ & IDF1(\%)↑ \\
\midrule
MeMOTR\cite{59}& 44.9& 53.3& 38.2& 62.1& 56.5\\
MOTIP\cite{39}& 46.6& 55.6& 39.4& 65.9& 57.8\\
 ByteTrack\cite{21}& 51.2& 59.0& 44.8& \textbf{71.3}&62.3\\
Hybrid-SORT-ReID\cite{60}& 49.4& 58.4& 42.0& 70.9& 59.8\\
Hybrid-SORT\cite{60}& 47.1& 56.8& 39.7& 68.4& 58.2\\
MOTR\cite{34}& 43.3& 48.7& 38.9& 55.3& 53.5\\
MOTRv2(baseline)\cite{11}& 49.1& 57.5& 42.3& 60.2& 53.7\\
\hline
VSD-MOT(ours) & \textbf{58.6}& \textbf{63.1}& \textbf{54.6}& 69.4& \textbf{65.2}\\
\bottomrule
\end{tabular}%
}
\end{table}
\begin{table}[htbp]
\centering
\caption{Performance Comparison of Different Trackers on the MOT Validation Set. ↑ Indicates Higher Is Better, the Best Results for Each Metric Are Highlighted in Bold. The Proposed VSD-MOT Outperforms Other Methods
on the MOT Validation Set. It Shows That after Training on the Mixed Training Set, the Proposed Method Can Maintain Good
Performance in Conventional Scenarios.}
\label{tab:4}
\Huge
\resizebox{\columnwidth}{!}{%
\begin{tabular}{c|ccccc}
\toprule
Tracker & HOTA(\%)↑ & DetA(\%)↑ & AssA(\%)↑ & MOTA(\%)↑ & IDF1(\%)↑ \\
\midrule
MeMOTR\cite{59}&  52.6& 59.9& 46.5& 67.4& 65.1\\
MOTIP\cite{39}& 54.3& 60.2& 49.3& 70.3& 67.5\\
 ByteTrack\cite{21}& 58.7& 64.4& 53.7& 76.7&71.5\\
Hybrid-SORT-ReID\cite{60}& 57.5& 64.2& 51.7& 76.6& 69.9\\
Hybrid-SORT\cite{60}& 55.7& 63.7& 49.1& 72.3& 67.8\\
MOTR\cite{34}& 48.2& 51.0& 45.8& 58.6& 59.6\\
MOTRv2(baseline)\cite{11}& 55.3& 61.3& 50.2& 66.4& 61.8\\
\hline
VSD-MOT(ours) & \textbf{70.8}& \textbf{72.6}& \textbf{69.0}& \textbf{77.8}& \textbf{76.6}\\
\bottomrule
\end{tabular}%
}
\end{table}

\subsection{Performance Comparison with the State-of-the-Art Methods on LQMOT and MOT}
We trained VSD-MOT and other state-of-the-art methods on the mixed training set of LQMOT and MOT with a ratio of low-quality videos to high-quality videos of 2:1, and tested them on the validation sets of LQMOT and MOT. The obtained results are presented in Table \ref{tab:3} and \ref{tab:4}. As can be seen from Table \ref{tab:3}, the proposed VSD-MOT outweighs other methods on the LQMOT validation set. Compared with other state-of-the-art methods, it leads by 3\% to 14\% in HOTA, DetA, and AssA, demonstrating the effectiveness of the proposed method in real-world low-quality video scenarios. As can be seen from Table \ref{tab:4}, the proposed VSD-MOT outperforms other methods on the MOT validation set. Compared with other state-of-the-art methods, it leads by 8\% to 21\% in HOTA, DetA, and AssA. It shows that after training on the mixed training set, the proposed method can maintain good performance in conventional scenarios.

\begin{table*}[t]
\centering
\caption{Effectiveness of Proposed Components on the LQMOT and MOT Validation Sets. \checkmark Represents the Baseline Equipped with the Corresponding Component. It Can Be Observed That Integrating
the Proposed Modules into the Baseline Gradually Enhances
Overall Performance}
\label{tab:5}
\Huge
\resizebox{0.9\textwidth}{!}{%
\begin{tabular}{c|c|cccc|ccccc}
\toprule
No.&  \multicolumn{1}{c|}{Dataset}& Baseline&CLIP Distillation&DCSD &DSWR& HOTA(\%)↑& DetA(\%)↑& AssA(\%)↑& MOTA(\%)↑&IDF1(\%)↑ \\
\midrule
1
&  & \checkmark&& && 49.1& 57.5& 42.3& 60.2& 53.7\\
 2& LQMOT& \checkmark& \checkmark&  && 52.7& 59.1& 47.2& 64.6& 59.5\\
 3& & \checkmark& \checkmark&\checkmark && 56.6& 62.3& 51.5& \textbf{69.7}& 63.2\\
 4& & \checkmark& \checkmark&\checkmark &\checkmark& \textbf{58.6}& \textbf{63.1}& \textbf{54.6}& 69.4& \textbf{65.2}\\
 \hline
 5& & \checkmark& & && 55.3& 61.3& 50.2& 66.4& 61.8\\
 6& MOT& \checkmark& \checkmark&  && 60.1& 64.1& 56.6& 71.5& 69.7\\
7&  & \checkmark&\checkmark&\checkmark && 67.9& 72.2& 63.9& \textbf{77.9}& 72.8\\
 8& & \checkmark& \checkmark&\checkmark &\checkmark& \textbf{70.8}& \textbf{72.6}& \textbf{69.0}& 77.8& \textbf{76.6}\\
 \bottomrule
\end{tabular}%
}
\end{table*}

\begin{table*}[t]
\centering
\caption{Comparison of Applying Different Ratio of High-quality to Low-quality Data in the Mixed Training Set  on the LQMOT and MOT Validation Sets. It Can Be Observed Better Results Are Achieved When the Ratio of High-quality Data to Low-quality Data Is 1:2.}
\label{tab:6}
\Huge
\resizebox{0.7\textwidth}{!}{%
\begin{tabular}{c|c|c|ccccc}
\toprule
 No.& Dataset&Low-quality：High-quality& HOTA(\%)↑ & DetA(\%)↑ & AssA(\%)↑ & MOTA(\%)↑ & IDF1(\%)↑ \\
\midrule
 1& &All High-quality& 48.9& 56.3& 42.6& 62.7& 55.2\\
 3& LQMOT&1:1& 55.3& 61.6& 49.8& 68.5& 62.9\\
 4& &2:1& \textbf{58.6}& \textbf{63.1}& \textbf{54.6}& \textbf{69.4}& \textbf{65.2}\\
 5& &All Low-quality& 54.1& 61.8& 47.5& 67.4& 62.6\\
\hline
 6& & All High-quality& 69.5& \textbf{78.4}& 61.8& \textbf{79.3}&71.7\\
 8& MOT& 1:1& 65.4& 70.4& 60.9&  76.8&72.3\\
 9& & 2:1&   \textbf{70.8}& 72.6& \textbf{69.0}& 77.8& \textbf{76.6}\\
 10& & All Low-quality& 62.3&  66.4& 58.6&  75.4&70.9\\
\bottomrule
\end{tabular}%
 
}
\end{table*}

\subsection{Ablation Study}
 \textit{1) Verification of Proposed Components:} This section provides an in-depth analysis of the impact of several key components in our approach, including the knowledge distillation of the frozen CLIP Image Encoder, and the DCSD method. For the ablation study, we adopt the MOTRv2 model as the baseline model. This baseline model is represented by No. 1 and No. 4 respectively in Table \ref{tab:5}. Table \ref{tab:5} summarizes the impacts of these components on the performance of the LQMOT and MOT validation sets, demonstrating consistent performance improvements on these datasets. As illustrated in Table \ref{tab:5}, the incremental incorporation of our proposed modules into the baseline model progressively enhances overall performance.

In Table \ref{tab:5}, No.2 and No.5represent the models augmented with knowledge distillation of the CLIP Image Encoder based on No.1 and No.4, respectively. No.2 and No.5 achieve comparable performance to No.1 and No.4 while reducing the number of parameters, indicating that knowledge distillation effectively retains the benefits of the CLIP Image Encoder with enhanced efficiency.
Subsequently, No.3 and No.6 correspond to the models with the addition of the DCSD method based on No.2 and No.5, respectively. Compared to No.2 and No.5, No.3 and No.6 show improvements across all performance metrics, suggesting that the DCSD method can more effectively transform global visual semantic information into features suitable for multi-object tracking tasks, thereby enhancing the algorithm's performance.

\textit{2) Analysis of Mixed Training Set:} In this section, we investigate the impact of the ratio of high-quality to low-quality data in the mixed training set on the algorithm's performance, and validate it on the validation sets of LQMOT and MOT in Table \ref{tab:6}. When trained exclusively on high-quality data, the proposed method exhibits suboptimal performance on the LQMOT validation set. In contrast, it demonstrates strong performance on the MOT validation set.​ At a 1:1 ratio of high-quality to low-quality data, the proposed method undergoes a substantial performance boost on the LQMOT validation set, with the MOTA score and the HOTA score improving by approximately 6\%. On the MOT validation set, there is a marginal decline of around 3\% to 4\% in both metrics.​ When the ratio of high-quality to low-quality data is adjusted to 1:2, both metrics show slight improvements across both validation sets.​ Training solely on low-quality data leads to performance degradation for the proposed method on both the LQDanceTrack and DanceTrack validation sets, yielding results inferior to those obtained with the mixed training set.​ These experimental findings indicate that, in comparison to training on purely high-quality or purely low-quality data, training on a mixed dataset enables the model to achieve superior performance in both low-quality scenarios and regular scenarios. Furthermore, the model trained with a 1:2 ratio of high-quality to low-quality data delivers the best overall performance.

\begin{table}[t]
\centering
\caption{Efficiency Analysis on the Impact of Introduced Components on Multi-Object Tracking Using the LQMOT and MOT Validation Sets. It Can Be Observed That an Alpha Value of 0.50 Results in the Best Overall Performance.}
\label{tab:7}
\Huge
\resizebox{\columnwidth}{!}{%
\begin{tabular}{c|c|ccccl}\toprule

 Dataset&Alpha& HOTA(\%)↑ &DetA(\%)↑ & AssA(\%)↑ & MOTA(\%)↑ &IDF1(\%)↑ \\\midrule

 &0.2& 53.7&58.1& 49.8& 67.3&61.8\\
 LQMOT&0.4& \textbf{58.6}& \textbf{63.1}& \textbf{54.6}& \textbf{69.4}& \textbf{65.2}\\ 
  &0.6& 54.5& 59.0& 50.5& 67.8&62.0\\
\hline
&0.2& 65.5&68.5& 62.8& 77.2&71.4\\
 MOT&0.4& \textbf{70.8}& \textbf{72.6}& \textbf{69.0}& \textbf{77.8}& \textbf{76.6}\\ 
  &0.6& 63.1& 65.5& 61.0& 76.0&70.5\\
 \bottomrule
\end{tabular}%
}
\end{table}

\begin{figure*}[htbp]
    \centering
    \includegraphics[width=1\linewidth]{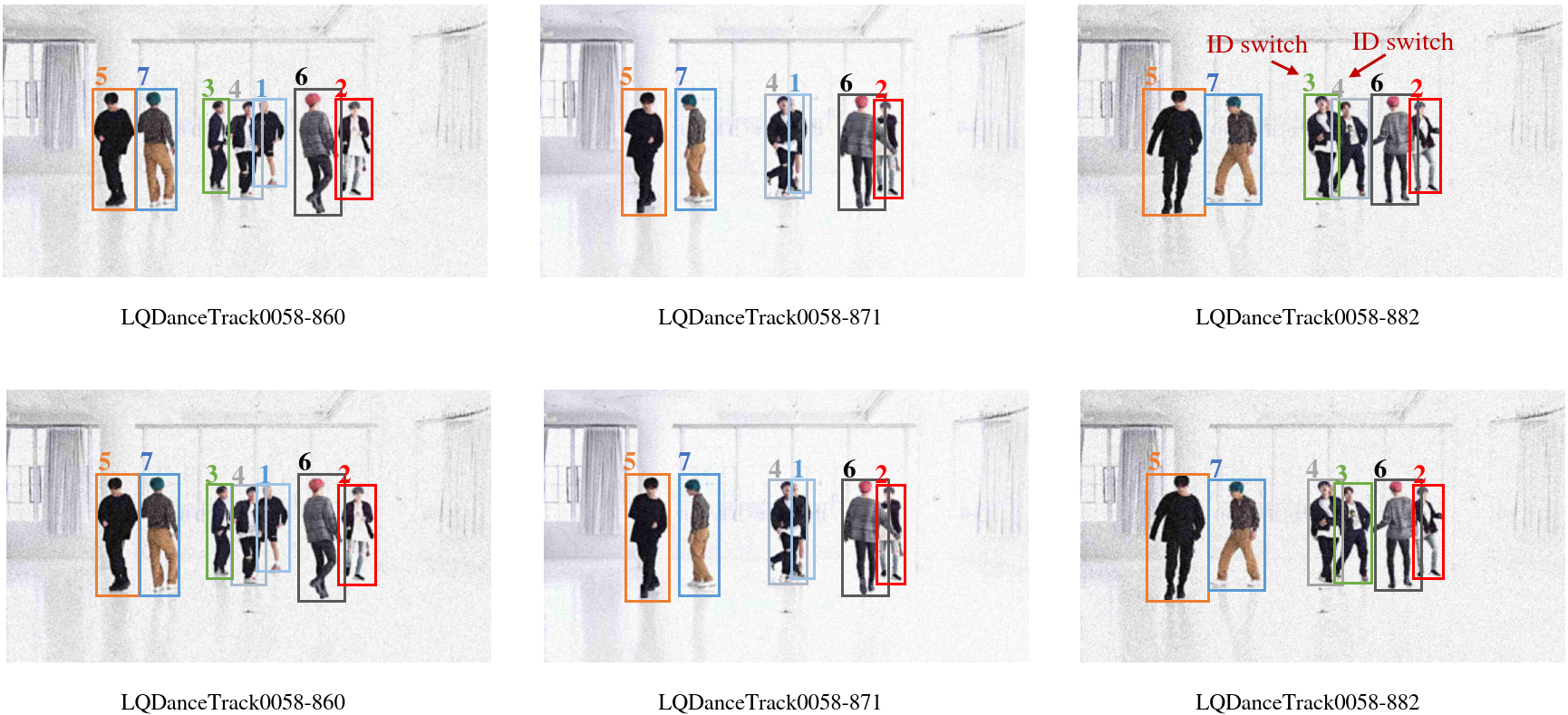}
    \caption{Partial results of the baseline method (first row) and SE-MOT (second row) on the LQDanceTrack validation set. In low-quality videos, our method achieves better tracking performance than the baseline model by introducing global visual semantic information of the images.}
    \label{fig:4}
\end{figure*}

\begin{figure*}[htbp]
    \centering
    \includegraphics[width=1\linewidth]{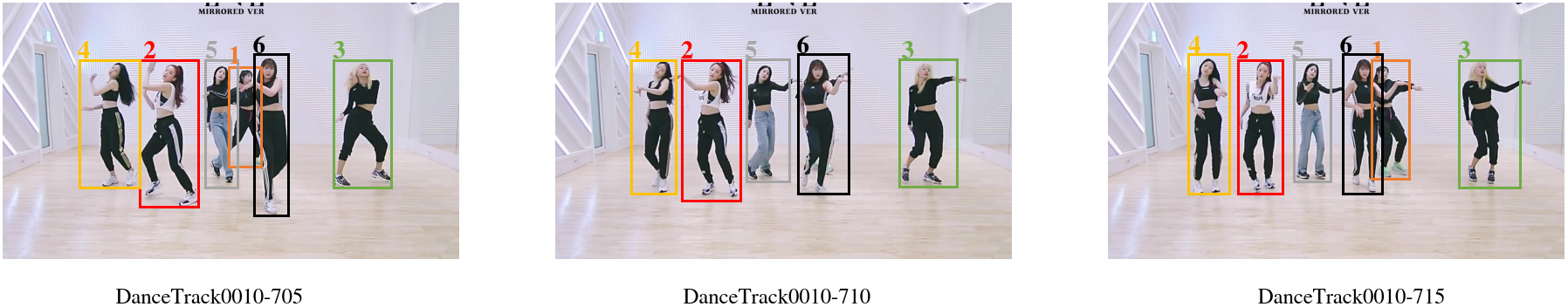}
    \caption{Partial results of SE-MOT on the DanceTrack validation set. Our method
can maintain good performance in conventional scenarios.}
    \label{fig:5}
\end{figure*}

\textit{3) Alpha Parameter Tuning:} The results for different alpha (α) values are shown in Table \ref{tab:7}. It can be clearly seen from the table that when the alpha value is 0.50, the overall performance is the best. Both lower alpha values (0.25) and higher alpha values (0.75) lead to performance degradation, which indicates that balanced weighting of the distillation loss is crucial for achieving optimal knowledge transfer.
\begin{table}[t]
\centering
\caption{Efficiency Analysis on the Impact of Introduced Components on Multi-Object Tracking Using the LQDanceTrack Validation Set. It can be observed that the introduction of the student model has no significant impact on the inference efficiency of the model.}
\label{tab:8}
\Huge
\resizebox{0.7\columnwidth}{!}{%
\begin{tabular}{c|ccc|cc}\toprule

No.& Baseline&Student Model&DSWR& Parameters (M)& FPS\\\midrule

1
& \checkmark&&&41.9& 15.8\\
2
& \checkmark&\checkmark&&42.6& 15.6\\ 
3
& \checkmark&\checkmark&\checkmark&42.6& 15.5\\
\bottomrule
\end{tabular}%
}
\end{table}

\textit{4) Efficiency Analysis:} In this section, we investigate the impact of introduced components in this paper on the efficiency of the multi-object tracking using the LQDanceTrack validation set in Table \ref{tab:8}. The student model and DSWR module involve only a small number of extra parameters and do not exert a noticeable impact on efficiency.

\subsection{Visualization}

To visually demonstrate the effectiveness of VSD-MOT in low-quality videos, we selected results from the low-quality multi-object tracking dataset LQDanceTrack. In this section,  we adopted the MOTRv2 model as the baseline model. Fig. \ref{fig:4} shows the visual outcomes of the baseline method (first row) and our method (second row). The distinct colors of the bounding boxes indicate unique identities within each image. In the images LQDanceTrack0058-882 (first row), targets No.3 and No.4 moved rapidly, and the poor image quality led to incorrect ID switching. To address such low-quality scenarios, our method incorporates the frozen CLIP Image Encoder, knowledge distillation, TFTH, and HAT-Adapter modules. The second row of Fig. \ref{fig:4} illustrates that our method can maintain accurate tracking in low-quality videos.

To visually demonstrate the effectiveness of VSD-MOT in conventional scenarios, we selected results from the DanceTrack dataset. Fig. \ref{fig:5} presents the visualization results of our method, with colored bounding boxes denoting distinct identities in each image. Fig. \ref{fig:5} shows that our method maintains good performance in conventional scenarios.

The visualizations highlight the effectiveness and superiority of our proposed method in real-world low-quality video scenarios, while also demonstrating robust performance in conventional scenes. This enhances the overall robustness of multi-object tracking algorithms across different scenarios.

\section{Conclusion}
This paper presents VSD-MOT, an end-to-end online tracking framework designed for simultaneous detection and tracking. To address the significant performance degradation of current multi-object tracking algorithms in low-quality video scenarios, we propose leveraging knowledge distillation to learn the ability of the teacher model, CLIP Image Encoder, to extract global visual semantic information from images, thereby compensating for the loss of information in low-quality images. Additionally, recognizing that the original visual semantic information extracted by the frozen CLIP Image Encoder is not well-suited for multi-object tracking tasks, we introduce the Dual-Constraint Semantic Distillation (DCSD) method to enable efficient knowledge transfer from teacher to student model. To address the dynamic variation of frame quality in low-quality videos, we propose the Dynamic Semantic Weight Regulation (DSWR) module, which adaptively allocates fusion weights based on real-time frame quality assessment through three main stages: Frame Quality Assessment, Weight Generation, and Adaptive Feature Fusion. The DSWR module follows the principle of "lower quality, higher semantic weight" to dynamically balance the fusion ratio between visual semantic features and query vector features, significantly enhancing tracking stability across varying frame quality conditions. Extensive experiments validate the efficacy of each component. Our method demonstrates superior performance in real-world low-quality video scenarios while maintaining robust performance in conventional settings.

\bibliography{VSD-MOT}
\bibliographystyle{IEEEtran}

\vfill
\end{CJK}
\end{document}